\documentclass[conference]{IEEEtran}
\usepackage{amsmath,amssymb,amsfonts}
\usepackage{algorithmic}
\usepackage{graphicx}
\usepackage{textcomp}
\usepackage{subfigure}
\usepackage{float}
\begin{document}
\title{Tool Breakage Detection using Deep Learning\\}


\author{
\textmd{\qquad\qquad\qquad\qquad\qquad\qquad\qquad}
\and
\textmd{\Large Guang Li\IEEEauthorrefmark{1}, Xin Yang\IEEEauthorrefmark{2}, Duanbing Chen\IEEEauthorrefmark{1}, Anxing Song\IEEEauthorrefmark{2}, Yuke Fang\IEEEauthorrefmark{2}, Junlin Zhou\IEEEauthorrefmark{1}}
\\
\and
\IEEEauthorblockA{
\IEEEauthorrefmark{1} UESTC, Big Data Research Center, Chengdu, China\\
leeguang923@gmail.com, \{dbchen, jlzhou\}@uestc.edu.cn\\
}
\and
\IEEEauthorblockA{
\IEEEauthorrefmark{2} Union Big Data, Chengdu, China\\
\{yangxin, fangyuke, songanxin\}@unionbigdata.com\\
}
}

\maketitle

\begin{abstract}
In manufacture, steel and other metals are mainly cut and shaped during the fabrication process by computer numerical control (CNC) machines.
To keep high productivity and efficiency of the fabrication process, 
engineers need to monitor the real-time process of CNC machines, 
and the lifetime management of machine tools. 
In a real manufacturing process, breakage of machine tools usually happens without any indication, this problem seriously affects the fabrication process for many years.
Previous studies suggested many different approaches for monitoring and detecting the breakage of machine tools.
However, there still exists a big gap between academic experiments and the complex real fabrication processes such as the high demands of real-time detections, the difficulty in data acquisition and transmission.
In this work, we use the spindle current approach to detect the breakage of machine tools, which has the high performance of real-time monitoring, low cost, and easy to install. 
We analyze the features of the current of a milling machine spindle through tools wearing processes, and then we predict the status of tool breakage by a convolutional neural network(CNN).
In addition, we use a BP neural network to understand the reliability of the CNN. 
The results show that our CNN approach can detect tool breakage with an accuracy of 93\%, 
while the best performance of BP is 80\%.
\end{abstract}


\section{Introduction}\label{sec:intro}
Detecting the tool breakage of CNC machines emerged in the actual industrial production has a great significance, 
tool breakage is often causing huge losses because people could not find the tool breakage in time. 
Manual detection of tool breakage is both time-consuming and laborious, 
it seriously affects the production efficiency and increases the cost of production. 
As a result, accurate detection of the tool breakage in a very short time will greatly enhance the production efficiency and reduce the costs of production. 
However, real-time detection of breakage tools is very difficult in the past. 

With the new techniques of cloud computing, big data, and machine learning, research on deep learning(e.g., neural network) is booming.
Many researchers start to explore the field of the industrial intelligence, especially the prediction of the tools wearing/breakage. 
During this period, many methods of detection proposed to predict the life of tools. 
Li analyzed the method that the tool breakage can be detected in the time-domain based the motor current signal \cite{Li2001Detection}. 
Youn introduced that cutting force of machine can be used to defect flank wear, the crater wear and the relationship between cutting force and cutting conditions \cite{Youn2001A}. 
Nie proposed a method that tools wear can be judged by acoustic emission signals with the wavelet packets technology and characteristic analysis \cite{Nie2009Application}. 
For the spindle current detection method, it has many advantages such as low cost, high performance on real-time, easy to install, and high precision. 
Based the advantages of current detection, many researchers judge tools wear by the current. 
Akbari studied the method that monitoring the tools wear by quantizing the harmonic distortion of spindle current and characteristic analysis in time and frequency domain \cite{Akbari2017A}. 
Lin defected the condition of tools through LS-SVM based on spindle current \cite{Lin2017Sequential}. 
Sevilla introduced the method of tool breakage detection based on a feed-motor current through the technology of discrete wavelet transform and statistical methodologies \cite{Sevilla2011Tool}. 
However, the current detection method is also difficult to monitor the tool breakage through the spindle current, because the changes of the wear reflected in the current are subtle, which causes the great troubles for monitoring tool breakage. 
With the development of technology, particularly in the field of artificial intelligence, neural networks are increasingly capable of extracting features and anomaly detection. 
And among neural networks, the performance of convolutional neural network (CNN) is especially prominent. 
Li estimated the feed cutting force by current and defected the tools wear based cutting force through fuzzy network-FNN 
\cite{Li2000Current}. 
Corne analyzed the effect of different neural networks with different neurons number based the current and the feasibility of prediction tools wear by neural network \cite{Corne2017Study} \cite{Corne2016Enhancing}. 

We found the difficulties of detecting tool breakage as follow: 
1. Experiments need high precision and time-sensitive data collection, because we need to precisely monitor real-time tool breakage. 
2. It is hard to extract the effective features can reflect tools wear.
3. The general method cannot capture effective information from a large amount of current data. 
To the best of our knowledge, Corne achieved a 94\% tool wear prediction with five neuron fuzzy neural networks and verify the feasibility of current approach by comparing to cutting force approach
\cite{Corne2016Enhancing}. 
We propose more complex networks will be more stable and reliable, so we plan to prove it by comparing CNN with five layers and BP with five neurons.
In actual processing, breakage of tools often unnaturally happens without any indication, we need the approach has the strong learning ability and robustness. 
In this work, we get the spindle current data by a high sampling rate Hall Sensor and NI data acquisition card. 
We performed time domain analysis of the preprocessed data and extracted satisfactory features. 
Considering the above problem, we choose CNN to achieve this work and BP as a comparison. 
Finally, CNN detects the tool breakage in a minute with the 93\% accuracy, the performance of our model is not good as Corne, 
but we found the performance of CNN is more robust and reliable than BP, and we believe the performance of CNN still has a lot of room for improvement.

The rest of this paper is organized as follows,
In Section \ref{sec:bg}, we talk about the technical background, which contains the main techniques used in the study.
In Section \ref{sec:approach}, we introduce the design of the experiment for detecting tool breakage. 
In Section \ref{sec:results}, we show our analysis results such as time domain features and the results of our network.
Section \ref{sec:conclusion} shows the conclusion part of this study.

\section{Background}\label{sec:bg}
\subsection{Convolutional neural network}
Within our knowledge, neural networks are favored for the excellent learning ability and fitting ability. Compared with neural networks, the main contributions of CNN as below:

\begin{itemize}
\item The CNN replaces the full connection with a local connection of the convolution kernels. The convolution kernels are composed of some parameters, which weights and averages the input data and slides with a set step size. Formulated as:
\begin{equation}
X=w \cdot x\
\end{equation}
where w is the convolution kernel matrix and x is the equal-sized input data block corresponding to the convolution kernels.

The convolution kernels function as a filter, for example, using a 3*3 convolution kernel with 8 in the middle and -1 around, deconvolving a picture shall play a role in sharpening. The parameters of convolution kernel can be understood as the preference of the network, for example, when we observe a person, we focus on her/his eyes, hair, nose, and height. Then the weight parameters of her/his eyes, hair, nose, and height may be higher than others.
\end{itemize}

\begin{itemize}
\item Weight sharing, to be simple, for a picture data, the same depth of the convolution kernel uses the same parameters to filter the picture data.For example, if our eyes are considered to be a convolution kernel, no matter what kind of image we see, we all use our eyes to feel it, and for an input data, the convolution kernel of the same depth is used regardless of where the data is convoluted. In this way, CNN saves a lot of parameters compared to traditional neural networks. At the same time, they also reduce overfitting.
\end{itemize}

\subsection{Backpropagation}
The derivative expression is as below:
\begin{equation}
\frac{df(x)}{dx} = \lim_{h->0}{\frac{f(x+h)-f(x)}{h}}
\end{equation}
where h is a very small value, close to 0, $\frac{d}{dx}$ is applied to function f, and return a derivative. The derivative could be seen as the slope of a line tangent to the function f, and the derivative indicates how sensitive the entire expression is to the value of the variable and the direction of the function drop. 

Backward propagation is the essence of neural networks, backprop recursively calculates expression gradients by chain rule. for example, an expression as below:
\begin{equation}
f(x,y,z) = (x+y)z^2
\end{equation}
It is a simple expression, we let u=x+y, the expression becomes f(x,y,z)=$uz^2$ and we can easily get the partial differential expression :
\begin{equation}
\begin{split}
\frac{\partial f}{\partial u} = z^2, & \frac{\partial f}{\partial z} = 2uz \\
\frac{\partial f}{\partial x} = \frac{\partial f}{\partial u} \cdot \frac{\partial u}{\partial x} = z^2*1, &
\frac{\partial f}{\partial y} = \frac{\partial f}{\partial u} \cdot \frac{\partial u}{\partial y} = z^2*1 
\end{split}
\end{equation}


In this way, the gradient of the nodes in the network can be quickly determined, the weights in the network will be updated according to the gradient, and the network will reach the best point faster.

\subsection{Normalization}
Normalization in two parts:
\begin{itemize}
\item Zero-mean, the formula is as follows :
\begin{equation}
x = x - \mu
\end{equation}

where the $\mu$ is the mean of the data.
\end{itemize}

\begin{itemize}
\item One-variance, the formula is as follows :
\begin{equation}
x = \frac{x}{\sigma^2}
\end{equation}

where the $\sigma$ is the variance of the data.
\end{itemize}

And combine the above two formula :
\begin{equation}
x = \frac{x-\mu}{\sigma^2}
\end{equation}

Why do we need to normalization and why do we need zero-mean and one-variance? Here is an example:

\begin{figure}[H]
\centering
\begin{minipage}{0.48\linewidth}
  \centerline{\includegraphics[width=4.8cm]{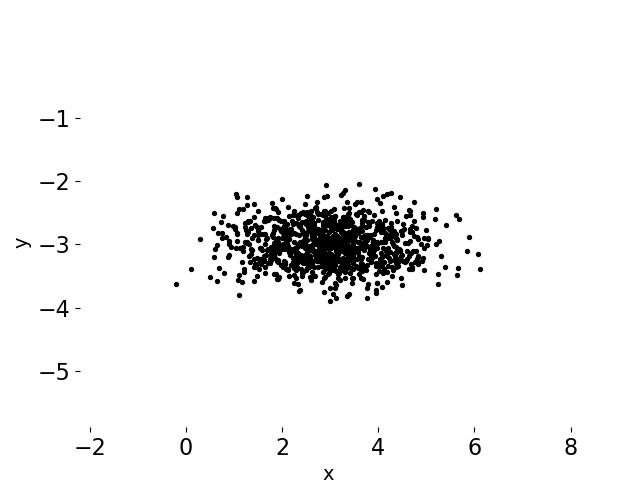}}
  \centerline{(a) original data}
\end{minipage}
\vfill
\centering
\begin{minipage}{0.48\linewidth}
  \centerline{\includegraphics[width=4.5cm]{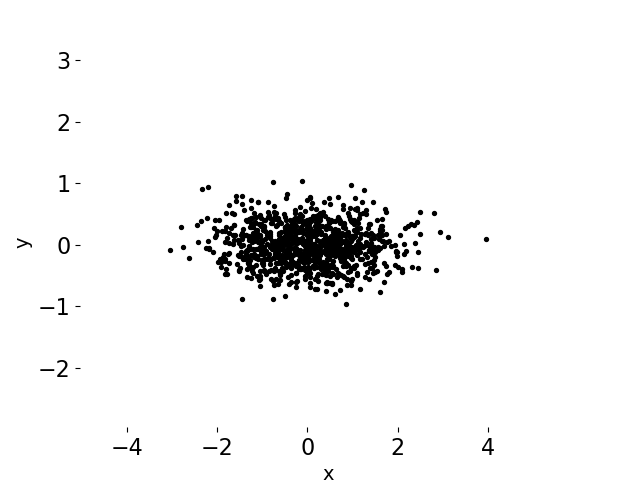}}
  \centerline{(b) Zero-mean value }
\end{minipage}
\hfill
\centering
\begin{minipage}{0.48\linewidth}
  \centerline{\includegraphics[width=4.5cm]{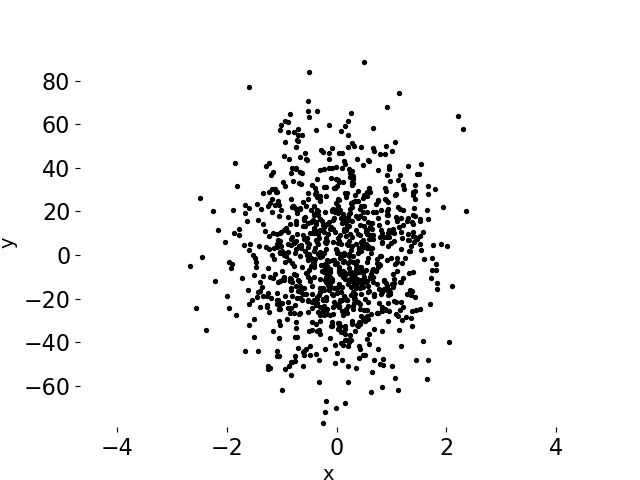}}
  \centerline{(c) Normalization value }
\end{minipage}
\hfill
\caption{Sample of Zero-mean and Normalization}
\label{fig:sample}
\end{figure}
In the above images, Fig \ref{fig:sample}(a) shows the randomly generated original data. Handled by zero-mean and we get Fig \ref{fig:sample}(b), we can observe that the data moved to the middle of the axis comparing the original data. Then, we handled the data of Fig \ref{fig:sample}(b) with driving $\sigma^2$ and get Fig \ref{fig:sample}(c). In Fig \ref{fig:sample}(c), the data in the x-axis and y-axis got balanced rather than the data length of one axis is more than the other axis. Therefore, data will be more symmetry and balanced by normalization. By normalization, we can close to the best point of network wherever the starting point is, and also we can turn up the learning rate to reach the best point faster. If the data is not symmetry and balanced, it could happen that count of data in an axis is more than the others, reaching the best point of a network will be very zigzagged even could not reach the best point and the learning rate of a network will be difficult to make a choice.

\section{Approach}\label{sec:approach}
In this work, we want to do: 1.Verify the signals of the spindle current can reflect tool wear. 2.Extract the features of tool wear. 3.Monitor tool breakage by CNN. We plan to check data and extract features before training network because a good data is more conducive to train the CNN and get a satisfying result. Meanwhile, features analysis and signal preprocessing can also be used as the experimental feasibility analysis. Although CNN has strong feature extraction and learning ability, if the data itself has problems or the current signal itself cannot reflect tool wear, CNN also cannot solve the problem. Therefore, it is necessary to perform feature analysis and signal preprocessing, we plan to do some analysis of the current signal from the perspective of the time domain, and then feed the selected feature into neural network we built, training and testing it.

\subsection{Experimental Environment and Data Description}
The data of this experiment was produced by a well-known manufactory with large-scale CNC machines, data is authentic and reliable, 
the current signal is detected according to the hall sensor installed on the spindle control wire.
During the data acquisition process, the data acquisition card perform the analog-to-digital conversions and processing steps like low-pass filtering. 
Finally, the data is sliced and stored in the buffer.

Different workpieces, tools, or even subtle changes in machine parameters may generate completely different data, which will also lead to different experimental results. Considering the experimental environment and machine model(Super MC F2.0-I/S), we plan to choose the cemented carbide milling tools to achieve this experiment. The most important parameters of this machine we must notice are as follows:

\begin{itemize}
\item Methods of processing, like Turning, Milling, Grinding, Drilling, Wire Cutting. Different processing methods produce different current signals. At the same time, the extent of wear on the tool is also different. In this experiment, we plan to use the Milling as our processing method.
\end{itemize}

\begin{itemize}
\item Rotating speed, the speed of the spindle directly affects the production accuracy of the workpiece and the tool wear. The speed is higher, the production efficiency is also enhanced, but the tool wear will more serious, and the different processing also requires different speeds. In this experiment, we used the speed of 6500 rpm for a machine.
\end{itemize}

\begin{itemize}
\item Feed rate, which means the relative displacement between the tool and the workpiece in the direction of feed motion. The feed rate affects the cutting cycle of the tool. Similar to the rotation speed, feed rate also directly affect the wear of the tool. Feed rate is larger, the tool wear will more severe. In this experiment, we plan to use the feed rate of 1500 mm/r.
\end{itemize}

Finally, considering the real-time requirements in actual processing, we plan to use Hall Current Sensor and NI PCIe data acquisition card to acquire the spindle current of the machine in this work, and the frequency of acquisition is 20000 current data points per second.

\subsection{Time Domain Feature Analysis }
Sensitive features of tool wear are more beneficial to neural network training and defecting the tool breakage, preprocessing of data is also more conducive to network convergence, such as normalization and denoising. We perform the feature analysis of current signals by some time domain analysis techniques before running a neural network.

Time domain analysis has intuitive and accurate advantages. Time domain can analyze the stability, transient and steady-state performance of a system, the time domain analysis methods commonly used are as follows :

\begin{itemize}
\item Mean, indicating the stability of the current over time and the center of signal change, it often gets mean value by the effective value of the current. In our experiment, we get mean value through the absolute value of the current to reduce the calculation, as:
\begin{equation}
\bar{x} = \frac{1}{N} \sum_{i=1}^{N}{\left|x_i\right|}
\end{equation}
\end{itemize}

\begin{itemize}
\item Root-mean-square, also known as the effective value, it can indicate the ability of the signal to send power, similar to mean, as :
\begin{equation}
X = \frac{1}{N} \sum_{i=1}^{N}{x_i^2}
\end{equation}
\end{itemize}

\begin{itemize}
\item Variance, expresses the extent of the sample deviates from the mean. It also reveals how much the sample fluctuates from one another, as :
\begin{equation}
\sigma^2 = \frac{1}{N} \sum_{i=1}^{N}{(x_i-\bar{x})^2}
\end{equation}
\end{itemize}

\begin{itemize}
\item Peak-to-peak value, measured the difference between the highest and lowest values of the signal in one cycle, which is the range between maximum and minimum. It also describes the size of the range of signal values, as :
\begin{equation}
X = max(x_i) - min(x_i)
\end{equation}
\end{itemize}

In our experiment, We perform time domain analysis with the time spans of 1 second, 1 minute and comparing time-domain analysis with actual machining process to observe whether the spindle current signal can reflect the tool wear and extract the signal features.

\subsection{The Architecture Of Neural Network }
The efficiency of CNN is related to the input data and various hyperparameters, some hyperparameters we need to explain are as follows:

\begin{itemize}
\item Depth of network, When the depth of the network is large, the capacity of the network is also large, and indicates that the network has the strong representational power to express more complex functions, but at the same time, the number of parameters in the network will be large, which means that the network has a huge amount of calculations and we need to take ways to prevent over-fitting. When the depth of the network is too small, it also leads to the representational power of the network is insufficient, even cannot fit data, not mention to get good results. So, the neural network is not good with large depth or too small depth. In our experiment, we constructed a network with the depth of five.
\end{itemize}

\begin{itemize}
\item Learning rate is also the most important hyperparameters of CNN. Take an example, if we compare the process of reaching the best point of the network to the downhill from the top of the mountain, the size of the learning rate could seem as the size of the steps we go downhill. If we take small steps, we need to spend a long time to reach the bottom of the mountain. If we take a big step, we may go to another slope, and it is equally difficult to reach the bottom of the mountain. So, a suitable learning rate can help us to reach the best point of the network faster. In our experiment, we set the value of learning rate is 0.0001.
\end{itemize}

\begin{itemize}
\item Weight initialization, a suitable weight will be more conducive to network convergence and gradient descent. In our experiment, we initialize weights with the normal distribution of 0 mean and 0.01 standard deviation.
\end{itemize}

The architecture we built (see Fig \ref{fig:CNN}):
\begin{figure}[htbp]
\centerline{\includegraphics[width=9cm,height=4.5cm]{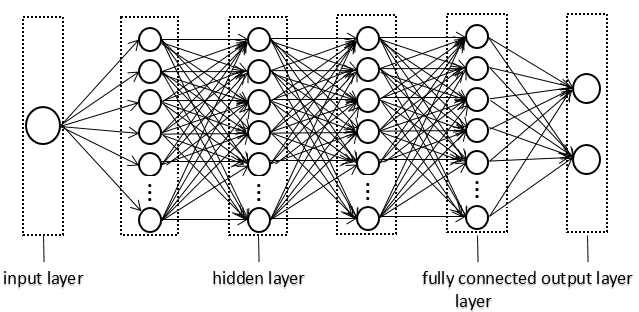}}
\caption{An illustration of the architecture of a CNN.}
\label{fig:CNN}
\end{figure}

In this work, the neural network we built includes an input layer, three hidden layers, and a fully connected layer. Finally, we get the output of prediction results through the softmax layer.
The data volume of the entire network is contained by the full machining cycle of five tools spindle current data, totals of 404 samples, 52080000 data points. Data is divided into training set and test set, the ratio of the training set and test set is 4:1, where the training set contains the current data of four tools, and the test set contains the full life cycle of one tool. In the input layer, the original data is randomly extracted from the training set,  and we handle it by normalization. Finally, we feed the data of size 1*7200*1 into the network. If the data does not reach 7200, zero-padding will be used. Deserve to be mentioned, in the training process, the input data is randomly selected, but in the test process, input data is extracted by the time sequence of tool processing. About the mini-batch of the network, the mini-batch size we selected is 20 during the training, but in the testing, mini-batch is 1. Finally, we set the label of normal data is 0, and the label of breakage tool data is 1.

The first convolutional layer in Fig \ref{fig:CNN} filters the [1*7200*1] input current feature data with 128 kernels of size [1*5*1] and a stride of 1, because we suggest that the kernels of [1*5*1] shall have the better performance to the kernels of [1*3*1] in our experiment, because the bigger kernels will acquire the more feature of the current wear. The second convolutional layer takes the output of the first hidden layer with the size of [1*3600*128] as inputs and filters it with 256 kernels of size [1*5*128]. The third convolutional layer has 512 kernels with the size of [1*5*256] and connected to the output of the second hidden layer with the size of [1*1800*256]. Then, we reshape the size of the output of third hidden layer with [1,900,512] to [1,1*900*512], the fully connected layer take it as the input and filter it with 1024 neural. Finally, we feed the output of fully connected layer into softmax layer and we get the output of the whole network with the size of [2]. In addition, each hidden layer contains a relu layer and a max-pooling layer with a kernel of size 1*2*1 and a stride of 2.

Although the size of our neural network is not big, the size of parameter is also closed to half billion, it is large and necessary to prevent network overfitting. Based on the good performance of overfitting and other aspects of batch normalization, we plan to add batch normalization layer behind the first convolutional layer, the second convolutional layer, and the third convolutional layer. A dropout layer is taken behind the third hidden layer to prevent overfitting.

To check the performance of the CNN, we built a BP neural network for comparison. The BP network is contained an input layer, a hidden layer with 512 neurons and an output layer with the size of 2. The input data, loss function,  and activation function is the same as CNN. The difference is that optimization of BP is Gradient Descent, and the optimization of CNN is Adam.

\section{Results}\label{sec:results}
In the time domain, we try to analyze the mean, variance, peak-to-peak of spindle current with a time span of 1 second, 1 minute to verify current signal can reflect tool wear in this way. Then, we use the feature of time domain as the input of CNN. Finally, We train the network described above and test it.
\subsection{Time domain analyze}
\begin{figure}[H]
\begin{minipage}{0.48\linewidth}
  \centerline{\includegraphics[width=4.0cm]{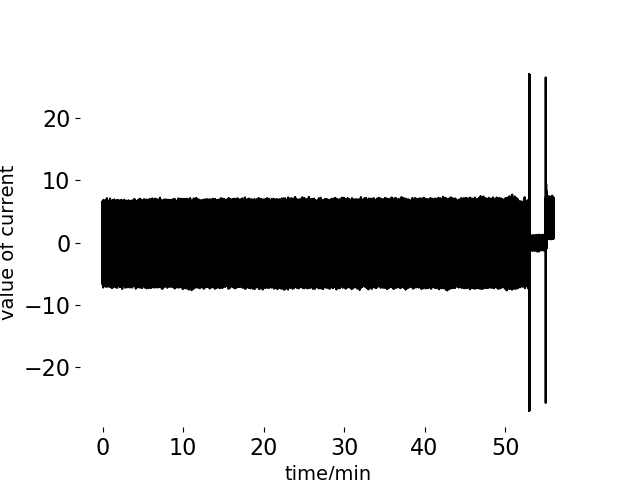}}
  \centerline{(a) original data}
\end{minipage}
\hfill
\begin{minipage}{.48\linewidth}
  \centerline{\includegraphics[width=4.0cm]{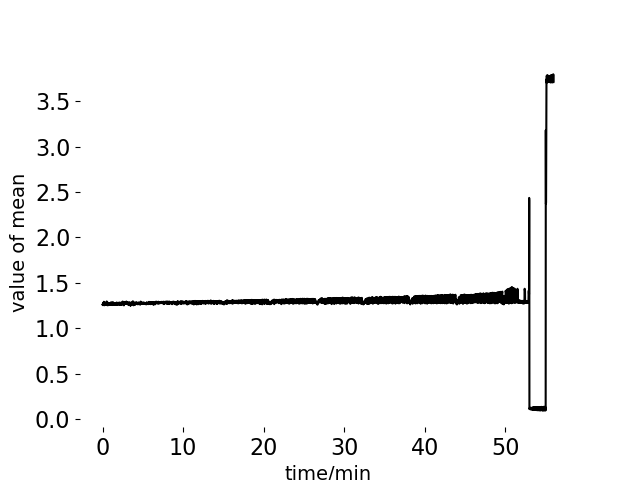}}
  \centerline{(b) mean value }
\end{minipage}
\vfill
\begin{minipage}{0.48\linewidth}
  \centerline{\includegraphics[width=4.0cm]{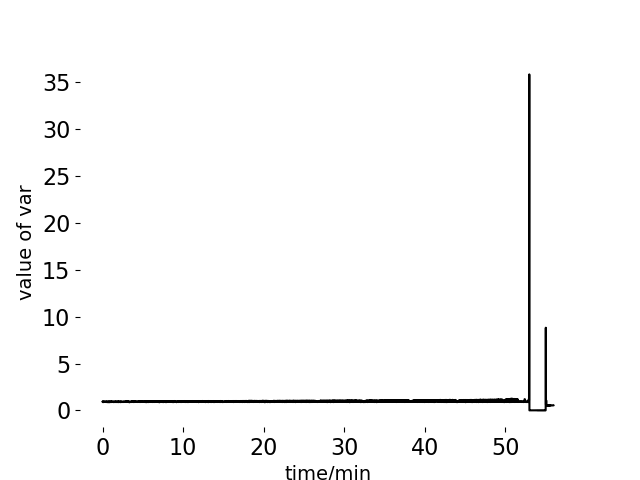}}
  \centerline{(c) variance value }
\end{minipage}
\hfill
\begin{minipage}{0.48\linewidth}
  \centerline{\includegraphics[width=4.0cm]{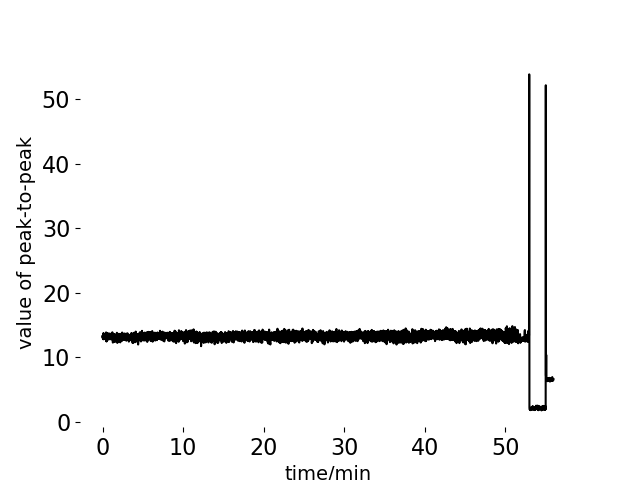}}
  \centerline{(d) peak to peak value}
\end{minipage}
\caption{Time domain analysis with time span of 1 second}
\label{fig:timedomain}
\end{figure}

we totally analyzed five tools and the results of these five tools are parallel. Both of these tools have a rising trend before the tool broken and have a dramatic change at the moment of tool breakage. we take an example of one tool in five tools as follow: 

After wiping off the unnecessary current data such as reset signal, we take the time domain analysis to the current data which contain a full life cycle of a tool (see Fig \ref{fig:timedomain}(a)). 
In Fig \ref{fig:timedomain}, we compared the original data with the data after time domain analysis. Even to original data in the picture, we can clearly observe the severe wear of tool at the last five minute. The trend of the current mean value is rise gradually in the starting of processing, and a big decline and rise are appeared at last (see Fig \ref{fig:timedomain}(b)). In addition, the trend of current variance value is similar to the current peak-to-peak value(see Fig \ref{fig:timedomain}(c) and Fig \ref{fig:timedomain}(d)). Both the variance and peak-to-peak can reveal the fluctuates extent of the signal, so, we can conclude that the current tool is windless in the earlier of the life of the tool and it explains the wear of the tool is smooth at the beginning, then, dramatic change appears at last of the tool life, Through repeated examination, the moment that the dramatic change appeared is corresponding to the time that the tool breakage was logged by worker(see Table \ref{tab:log}). 

The processing log (Table \ref{tab:log}) show that the life cycle of the tool is fifty-five minute and the cutting sound size of the machine. In the actual processing environment, the machine will not be stopped until the tool is broken or the machine is abnormal. so, the worker of processing center only to judge the breakage of tools by the sound size of a machine by experience. But they will check whether the tool is really broken when the sound of the machine is abnormal. In fact, it is later than the actual moment of the breakage of a tool when the worker judge the tool is broken, and the worker only judges a period of the breakage of a tool. Comparing Fig \ref{fig:timedomain} and Table \ref{tab:log}, it can be observed that the dramatic change of the current signal appear at 52-55 minutes and it is judged as breakage tool at 46-55 minutes in the table of the processing log. But this has little effect, as long as we are sure that the tool is broken during this time and we check it by the current signal.

\begin{table}[htbp]
\caption{Table of Processing log}
\centering
\begin{tabular}{rrl}
\hline 
Start\;Time&End\;Time&Cutting\;sound\;size\\
\hline  
0&12&slight\\
\hline 
12&35&normal\\
\hline
35&46&slightly larger\\
\hline
46&55&abnormal\\
\hline
\end{tabular}
\label{tab:log}
\end{table}

\begin{figure}[H]
\centering
\begin{minipage}{.48\linewidth}
  \centerline{\includegraphics[width=5.0cm]{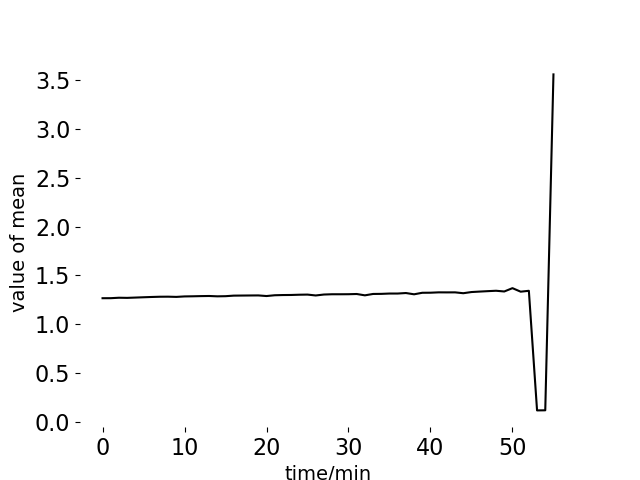}}
  \centerline{(a) mean value }
\end{minipage}
\vfill
\begin{minipage}{0.48\linewidth}
  \centerline{\includegraphics[width=4.5cm]{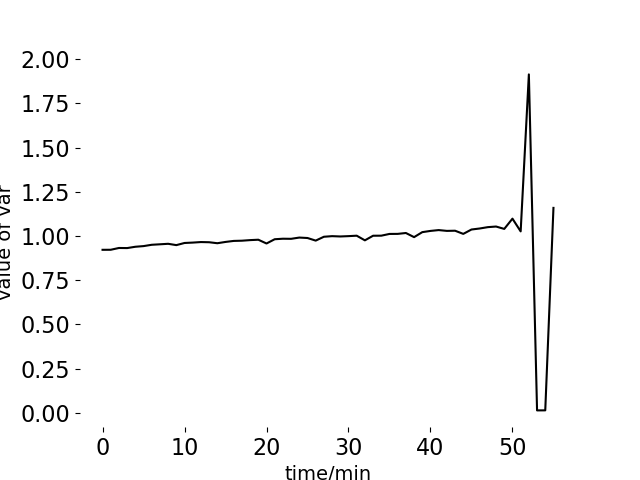}}
  \centerline{(b) variance value }
\end{minipage}
\hfill
\begin{minipage}{0.48\linewidth}
  \centerline{\includegraphics[width=4.5cm]{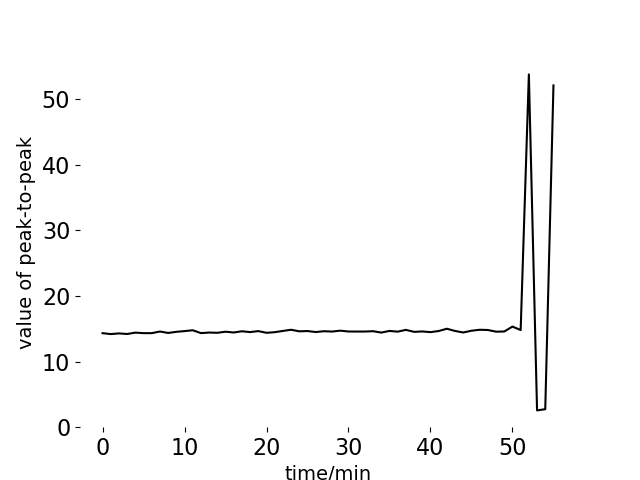}}
  \centerline{(c) peak to peak value}
\end{minipage}
\caption{Time domain analysis with time span of one minute}
\label{fig:1min}
\end{figure}

In addition to the time domain analysis of the current signal in a second, we also do the experiment of the time domain analysis in a minute(see Fig \ref{fig:1min}). Comparing Fig \ref{fig:timedomain}, the one-minute time domain feature is more smooth. It is easy to understand that increased time span is like smoothing the current signal.

The result of above analysis verified that the tool wear can be reflected through the spindle current, the features of time domain can also reflect tool breakage, and the method that predicts tool breakage based current is feasible, we plan to take the mean value as the input of CNN and test it in the next step.

\subsection{Tool Breakage Prediction by BP and CNN}
We entered our model in the mean feature of the 4 tools, and run it no less than five times, results of BP is fluctuating, we report the average results in Table \ref{tab:performance}:

\begin{table}[h]
\caption{Performance of CNN and BP}
\centering
\begin{tabular}{lrr}
\hline 
Model&BP&CNN\\
\hline  
Number\;of\;iterations&2400&2400\\
\hline 
Size\;of\;mini-batch&20&20\\
\hline
Training\;accuracy&0.7&1\\
\hline
Validation\;accuracy&0.77&0.92\\
\hline
Testing\;accuracy&0.8&0.93\\
\hline
\end{tabular}
\label{tab:performance}
\end{table}

\begin{figure}[H]
\centering
\begin{minipage}{0.48\linewidth}
  \centerline{\includegraphics[width=4.5cm]{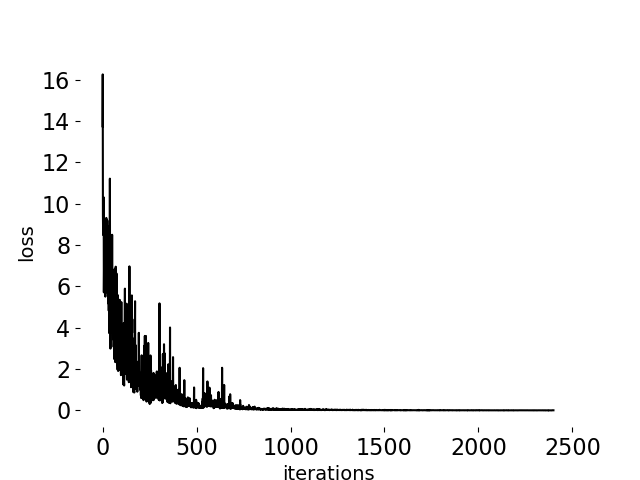}}
  \centerline{(a) Loss value of CNN}
\end{minipage}
\hfill
\centering
\begin{minipage}{0.48\linewidth}
  \centerline{\includegraphics[width=4.5cm]{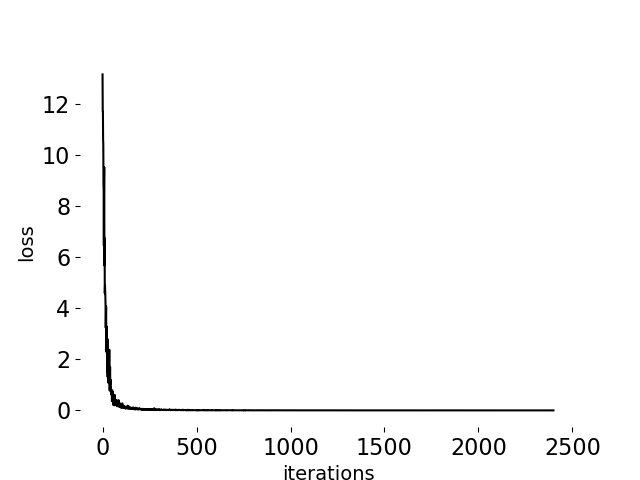}}
  \centerline{(b) Loss value of BP }
\end{minipage}
\caption{Loss value of the neural network}
\label{fig:loss}
\end{figure}

The best performance achieved by BP in this experiment was 80\%, bp is not stable, its training accuracy often has more than 80\% and less than 60\%, and the same as the testing accuracy. CNN is stable, the training accuracy of CNN is fixed at 100\%, and the testing accuracy is stable at more than 93\%. It may be due to CNN is more complex and the number of weight is more than bp. BP also has advantages such as the loss of BP is more smooth than CNN and faster convergence(see Fig \ref{fig:loss}). 

The performance of the CNN is satisfactory. The test data of network is the mean value of a minute, it means that the network can check the breakage tool in the one minute and give the warning to the worker and the worker can exchange the tool of the machine. From the accuracy of the neural network, it also indicates that the robustness of CNN is better than BP, and the predicted result of CNN is more reliable. It also shows that the performance of multilayer neural network CNN is really better than neural network with one-hidden layer BP.

\section{Conclusion}\label{sec:conclusion}
Prediction of tool breakage and the wear of tools is a combination of artificial intelligence and industry, it has enormous economic value and research value. In our experiment, feature engineering and deep learning are applied for real-time monitoring tool breakage, we verified the feasibility of current detection method and detected the tool breakage through the CNN and BP network with one hidden layer. Multilayer neural network worked better than the BP neural network with single hidden layer by comparing the results of BP and CNN. 
We have some deficiencies and improvement points:

\begin{itemize}
\item Insufficient sample proportion. In our sample, the positive sample accounts for 90\% of the total sample, because the duration of severe wear of tools to tool breakage is short. In this case, our network is good at testing the positive sampes and has general performance on the negative sampes. In fact, the remaining 7\% test error rate is mainly due to the network is not enough to fit negative sample.
\end{itemize}

\begin{itemize}
\item We get the prediction result from each minute by testing step of CNN. But in the actual situation, Real-time ask us to decrease the time of detecting tool breakage, because the time we spent is not only the operation of the neural network but the time of data collection, transmission, denoising and so on. So, we plan to improve our neural network and input data to achieve detecting the tool breakage in one second.
\end{itemize}

\begin{itemize}
\item Results in this paper are built in the fixed machine, fixed parameter, fixed tools, and materials. If this environment is changed, the results will be different, because the wear of tools is affected to a large extent by this environment element. We will continue to study the causes of tool chipping and prediction of  tool wear. We plan to take some novel methods on neural networks in this study and look forward to better results.
\end{itemize}

\bibliographystyle{IEEEtran}
\bibliography{IEEEabrv,paper_ref}
    
\end{document}